\documentclass[letterpaper]{article} 
\usepackage[preprint]{aaai2027}  
\usepackage[hyphens]{url}  
\usepackage{graphicx} 
\usepackage{amsmath}
\usepackage{natbib}  
\usepackage{caption} 
\usepackage{algorithm}
\usepackage{algorithmic}

\usepackage{newfloat}
\usepackage{listings}
\DeclareCaptionStyle{ruled}{labelfont=normalfont,labelsep=colon,strut=off} 
\floatstyle{ruled}
\newfloat{listing}{tb}{lst}{}
\floatname{listing}{Listing}

\usepackage{booktabs}
\usepackage{multirow}

\title{DARTree: Speculative Diffusion Decoding with Autoregressive Draft Trees}

\author {
    Tianyi Li,
    Yaxin Luo,
    Xinyi Shang,
    Zhiqiang Shen
}
\affiliations {
    VILA Lab, MBZUAI\\
    \{Tianyi.Li, Yaxin.Luo, Xinyi.Shang, Zhiqiang.Shen\}@mbzuai.ac.ae\\
    [0.35em]
    \textbf{Code: }{\color{magenta}\url{https://github.com/VILA-Lab/DARTree}}
}

\begin{document}

\maketitle

\begin{abstract}

Speculative decoding losslessly accelerates autoregressive language models by verifying multiple draft tokens in parallel. Diffusion-based drafters further reduce proposal latency by predicting an entire token block in parallel, but their position-wise distributions are marginal rather than conditioned on tokens selected along each draft path. Existing recurrent correction incorporates causal information along a single draft chain, whereas diffusion-based tree construction broadens candidate coverage without carrying this correction along individual branches. We introduce DARTree, a training-free speculative decoding method that extends a pretrained AR correction head from chains to trees. DARTree first constructs a fixed-width candidate tree by expanding and scoring all nodes at each depth in a single batch, and then only applies best-first pruning to select the verification tree, decoupling AR-head inference from sequential heap operations. Across seven math, code, and chat benchmarks, DARTree achieves the highest average acceptance length and speedup in all four model--temperature configurations, accepting up to 12.97 tokens per verification round, 98.6\% more than DFlash and 27.9\% more than Domino in the same setting, and reaching up to 9.73$\times$ lossless speedup over locally measured autoregressive decoding. 
\end{abstract}


\section{Introduction}
\label{sec:intro}
While modern autoregressive (AR) large language models (LLMs) have achieved remarkable success in understanding, reasoning, and generating text and code~\cite{achiam2023gpt,guo2025deepseek}, their inherently sequential nature limits inference speed, resulting in noticeable latency for users. 
At each decoding step, the model generates only one token per sequence while requiring access to nearly all model parameters, making inference predominantly I/O-bound and wasting the parallel computation capabilities of modern hardware.

To accelerate inference and improve hardware utilization, Speculative Decoding~\cite{leviathan2023fast,chen2023accelerating}
uses a lightweight drafter to propose several future tokens and verifies them in parallel with the target model, preserving the target model's output distribution. However, autoregressive drafters still produce proposals sequentially, so their latency grows with draft length. Diffusion-based block drafters remove this rollout by predicting an entire block in one forward pass, simultaneously lowering proposal latency and enabling higher-capacity drafters that improve proposal quality and acceptance, as demonstrated by DFlash~\cite{chen2026dflash}.

\begin{figure}[t]
    \centering
    \makebox[\columnwidth][r]{%
        \includegraphics[width=1.06\columnwidth,trim=8 6 8 5,clip]{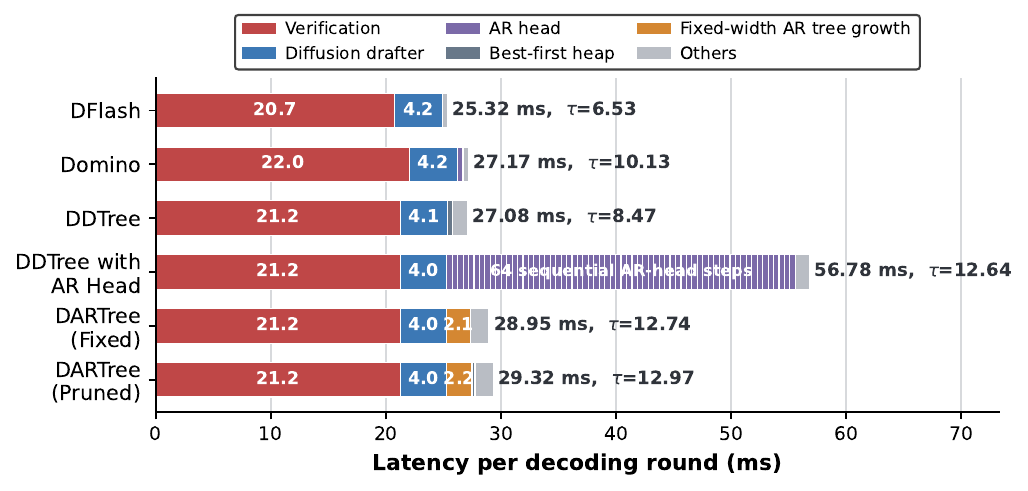}%
    }
    \par
    \includegraphics[width=\columnwidth]{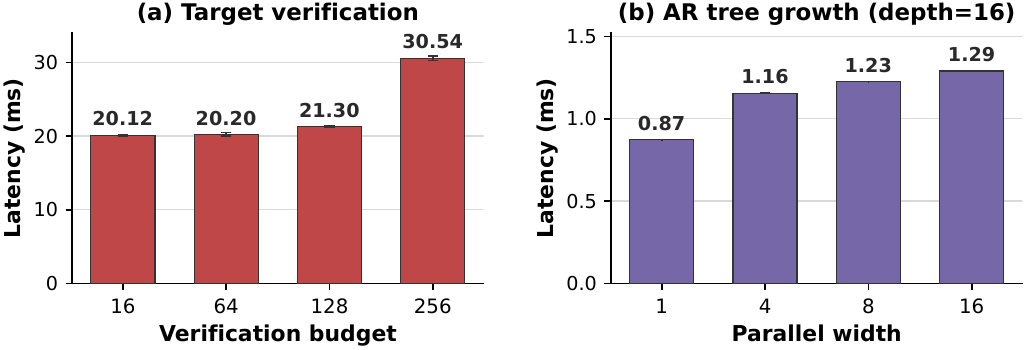}
    \caption{Latency breakdown and effects of verification budget and parallel width on GSM8K (Qwen3-4B, $T=0$). DARTree avoids sequential AR-head expansion and scales efficiently with budget and parallel width.}
    \label{fig:dartree_breakdown}
    \label{fig:dartree_budget_parallelism_latency}
\end{figure}

This parallelism comes at the cost of causal conditioning. Each draft position is predicted from the verified prefix but not from the realized tokens at earlier positions in the same block, creating a mismatch with the target model's conditioned distributions. Lightweight causal correction heads mitigate this mismatch by propagating selected tokens along the draft chain, partially restoring the missing intra-block dependencies without running the parallel backbone~\cite{huang2026domino,cheng2026dspark}. 

\begin{figure*}[t]
    \centering
    \includegraphics[width=\textwidth]{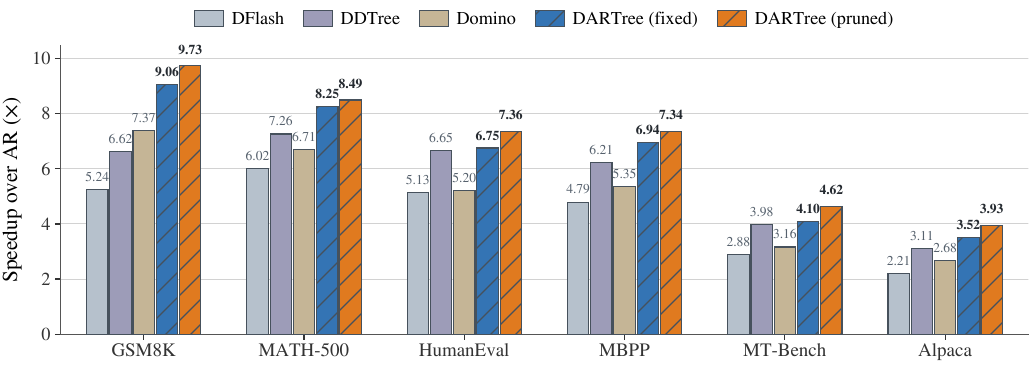}
    \caption{Speculative decoding speedup over standard autoregressive decoding across six benchmarks. DARTree with fixed-width construction and subsequent pruning consistently outperforms prior diffusion-based speculative decoding methods.}
    \label{fig:dartree_speedup_comparison}
    \vspace{-5pt}
\end{figure*}

Draft-tree verification provides a separate way to improve speculative decoding by exposing multiple candidate continuations to the target model. A prefix tree can be scored in one forward pass using tree attention, improving candidate coverage when an early draft position is uncertain. DDTree~\cite{ringel2026accelerating} constructs such a tree from the position-wise distributions of a block-parallel diffusion drafter; under a fixed node budget, its best-first search selects the highest-mass prefixes to maximize the surrogate expected acceptance length with a max heap. Draft trees and causal correction therefore address complementary limitations: the former broadens candidate coverage, while the latter improves causal dependency across draft positions.

Combining causal correction with this heap-based construction, however, creates a new efficiency bottleneck. Because the corrected scores of a node's children depend on that node's path-specific state, each heap pop must be followed by correction-head inference and subsequent heap pushes before the next node can be expanded. This interleaving forces correction-head inference to proceed one node at a time, making tree construction a substantial source of latency, as shown in Figure~\ref{fig:dartree_breakdown}. The key challenge is to retain path-dependent scoring while batching multiple branches.

We introduce \textbf{DARTree}, a training-free speculative decoding method that extends a pretrained causally corrected block-parallel drafter from a single chain to a speculative tree. DARTree expands a fixed number of nodes at each depth and evaluates their corrected distributions in one batch, avoiding the node-wise correction interleaved with heap operations. Since wider layer expansion adds little latency, DARTree first constructs a wider supertree and then applies best-first pruning later to obtain a compact verification tree, decoupling correction-head inference from sequential heap operations.

Across seven math, code, and chat benchmarks with Qwen3-4B and Qwen3-8B at temperatures 0 and 1, DARTree achieves the highest overall average acceptance length and speedup in all four model--temperature configurations. Figure~\ref{fig:dartree_speedup_comparison} summarizes the decoding speedups across representative benchmarks. 
Our contributions are threefold:
\begin{itemize}
    \item We identify the sequential bottleneck caused by coupling path-conditioned causal correction with node-wise best-first tree construction.
    \item We introduce a training-free, depth-wise batched tree-construction method that applies causal correction across multiple branches in parallel, achieving lossless speedup.
    \item We propose a deferred and simplified best-first pruning strategy for selecting the final verification tree.
\end{itemize}
\section{Related Work}
\label{sec:related}

\paragraph{Speculative Decoding.}
Speculative decoding accelerates autoregressive generation by drafting multiple future tokens and verifying them in parallel through the target model with rejection sampling~\cite{leviathan2023fast,xia2023speculative,chen2023accelerating}. Standard multi-token prediction (MTP) can serve as a lightweight drafting mechanism, using auxiliary prediction heads to propose tokens at multiple future positions in a single forward pass~\cite{gloeckle2024better}. Medusa~\cite{cai2024medusa} augments the target model with multiple prediction heads and organizes their candidates into a tree for parallel verification. The EAGLE family instead employs a learned drafter conditioned on internal features of the target model. EAGLE~\cite{li2024eagle} autoregressively predicts the target model's hidden states and maps them to candidate-token distributions using the target model's output head; EAGLE-2~\cite{li2024eagle2} dynamically allocates draft-tree nodes under a fixed node budget using context-dependent confidence scores; and EAGLE-3~\cite{li2025eagle3} replaces feature prediction with direct token prediction using fused multi-layer target-model features. 

\paragraph{Diffusion Language Models.}
Diffusion language models (DLMs) generate sequences through iterative denoising and update multiple tokens in parallel, offering an alternative to left-to-right autoregressive generation~\cite{austin2021structured,nie2025largelanguagediffusionmodels,ye2025dream,li2025survey,myrzakhan2026sink,gong2022diffuseq}. Recent masked DLMs such as LLaDA~\cite{nie2025largelanguagediffusionmodels} demonstrate that this paradigm can scale to large language models, while block diffusion generates autoregressively across blocks and denoises tokens within each block in parallel, thereby recovering compatibility with KV caching~\cite{arriola2025block}.
However, the parallelism of DLMs introduces a fundamental speed-quality trade-off: accepting more tokens per denoising step improves throughput but often degrades generation quality, whereas conservative decoding requires many refinement steps~\cite{li2025survey,wu2025fast}. Moreover, repeated bidirectional computation makes standard KV caching difficult and leads to unfavorable scaling on long sequences. These limitations motivate using DLMs as efficient draft generators under autoregressive verification, rather than as standalone replacements for autoregressive models.

\paragraph{Diffusion-Based Parallel Drafting.}
Diffusion-based drafting exploits sequence-level parallelism to reduce proposal latency~\cite{christopher2025speculative}. DFlash~\cite{chen2026dflash} introduces a lightweight block drafter that generates an entire proposal block in a single forward pass, achieving strong drafting quality and acceleration.
Plain one-pass diffusion-based drafters ignore intra-block causal dependencies; Domino~\cite{huang2026domino} and DSpark~\cite{cheng2026dspark} partially recover them with lightweight causal correction heads.
A complementary line organizes parallel draft predictions into trees. DDTree~\cite{ringel2026accelerating} derives a globally optimal prefix tree under a fixed node budget with respect to its diffusion-marginal surrogate, using best-first heap expansion.
In its primary Domino instantiation, DARTree integrates these directions by extending full-prefix AR correction from a single draft chain to multiple tree branches. It expands and scores candidate branches depth-wise in parallel and, after completing the candidate supertree, applies a single global top-$B$ selection to form the verification tree, avoiding sequential heap operations during correction.
\section{Preliminaries}
\label{sec:pre}

\subsection{Speculative Sampling}

Let $p$ and $q$ denote the target and draft models, respectively, and let $x_{1:t}$ be the tokens committed before a speculative decoding round. The drafter proposes $\gamma$ tokens $y_{1:\gamma}$, where $y_{<i}=(y_1,\ldots,y_{i-1})$ and $c_i=(x_{1:t},y_{<i})$ is the context at draft position $i$. When $T=0$, verification accepts consecutive draft tokens that match the target model's greedy predictions. At the first mismatch, it emits the target prediction and terminates the current round.

When $T>0$, let $p_T$ and $q_T$ be the temperature-scaled target and draft distributions. For $y_i\sim q_T(\cdot\mid c_i)$, standard speculative sampling accepts $y_i$ with probability
\begin{equation}
    a_i=\min\!\left(1,\frac{p_T(y_i\mid c_i)}{q_T(y_i\mid c_i)}\right).
\end{equation}
If accepted, $y_i$ is committed and verification continues to position $i+1$. At the first rejection, only one replacement token is sampled from
\begin{equation}
    r_T(v\mid c_i)
    =\frac{[p_T(v\mid c_i)-q_T(v\mid c_i)]_+}
    {\sum_{w\in\mathcal{V}}[p_T(w\mid c_i)-q_T(w\mid c_i)]_+},
\end{equation}
where $v$ denotes a possible replacement token, $w$ indexes the vocabulary $\mathcal{V}$ for normalization, and $[z]_+=\max(z,0)$. Every round commits the accepted draft prefix followed by one additional output token as a bonus. At the first rejection, this token is sampled from $r_T$, the remaining proposals are discarded, and the round ends; if all $\gamma$ proposals are accepted, it is sampled directly from $p_T$. In either case, the committed sequence follows the target distribution~\cite{leviathan2023fast,chen2023accelerating}.

This rejection-sampling procedure is not used by all implementations. DFlash~\cite{chen2026dflash} instead performs simple exact target-sample matching: it accepts consecutive draft tokens while they match the corresponding target samples, emits the target sample at the first mismatch, and terminates the round. This verification procedure preserves the target distribution without explicitly computing draft--target probability ratios or residual corrections.

\begin{figure*}[!t]
    \centering
    \includegraphics[width=\textwidth]{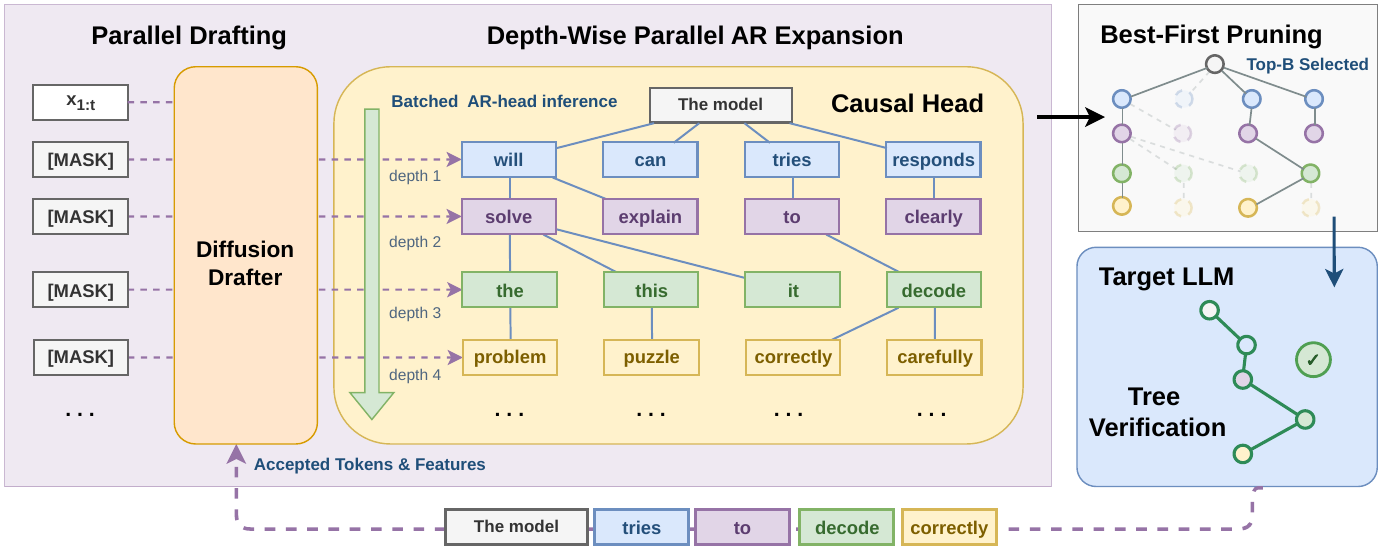}
    \caption{\textbf{Overview of DARTree.} Block-parallel drafting is followed by depth-wise batched AR expansion, deferred best-first pruning, and a single standard target-model tree-verification pass. Nodes sharing the same color are generated in the same depth-wise AR-head batch; rounded boxes indicate parameterized modules, while faded nodes are discarded by top-$B$ pruning.}
    \label{fig:method-overview}
    \vspace{-5pt}
\end{figure*}

\subsection{Parallel Drafting and Causal Correction}

Block-parallel drafters predict the distributions of all $\gamma$ future positions in one forward pass. Given a verified prefix $x_{1:t}$, these predictions induce the factorized distribution
\begin{equation}
    q_{\mathrm{par}}(y_{1:\gamma}\mid x_{1:t})
    = \prod_{i=1}^{\gamma} q_i(y_i\mid x_{1:t}).
\end{equation}
By eliminating sequential draft rollout, block-parallel drafting not only reduces proposal latency but also accommodates a higher-capacity drafter, thereby improving proposal quality and acceptance, as demonstrated by DFlash~\cite{chen2026dflash}. However, each $q_i$ conditions only on $x_{1:t}$, whereas the target distribution $p(y_i\mid x_{1:t},y_{<i})$ also depends on the realized earlier draft tokens. This missing intra-block dependency is the causal mismatch of parallel drafting.

To mitigate this mismatch, causal correction introduces dependencies among parallel predictions. Markov correction models the one-step transition from $y_{i-1}$ to $y_i$~\cite{cheng2026dspark}, which can also be approximated in parallel~\cite{rheinboldt2026treeflash}. RNN correction instead propagates a recurrent state $r_i$ that summarizes the realized prefix $y_{<i}$~\cite{huang2026domino,cheng2026dspark}.

\subsection{Tree Verification and DDTree}

A draft tree is a prefix tree that represents multiple candidate continuations. For verification, its nodes are flattened, and a tree-attention mask allows each node to attend only to the verified context and the tokens on its root-to-node path. The target model can therefore score all branches in one forward pass, after which the accepted root-to-leaf prefix is selected according to the target model.

DDTree~\cite{ringel2026accelerating} constructs a tree of at most $B$ nodes by maximizing a surrogate expected acceptance length under the factorized block-diffusion distribution $q_{\mathrm{par}}$ defined above. DDTree optimizes
\begin{equation}
    \max_{\mathcal{T}:|\mathcal{T}|\leq B}
    \mathrm{E}_{Y_{1:\gamma}\sim q_{\mathrm{par}}(\cdot\mid x_{1:t})}
    \!\left[\alpha_{\mathcal{T}}(Y_{1:\gamma})\right].
\end{equation}
For a concrete continuation $y_{1:\gamma}$, its acceptance length is $\alpha_{\mathcal{T}}(y_{1:\gamma})=\max\{d:y_{1:d}\in\mathcal{T}\}$, and $\alpha_{\mathcal{T}}(Y_{1:\gamma})$ is the random length of the longest prefix of $Y_{1:\gamma}$ contained in the tree. For a prefix $u=(u_1,\ldots,u_d)$, its mass is $q_{\mathrm{par}}(u\mid x_{1:t})=\prod\nolimits_{i=1}^{d}q_i(u_i\mid x_{1:t})$. The expectation above then decomposes into $\sum\nolimits_{u\in\mathcal{T}}q_{\mathrm{par}}(u\mid x_{1:t})$. Since an ancestor has no smaller mass than its descendants, the top-$B$ prefixes with the highest probability mass form the optimal tree under this surrogate. DDTree enumerates them with best-first search using a max-heap: each pop selects the highest-mass prefix and pushes its next sibling and first (highest-probability) child, yielding $O(B\log B)$ complexity for constructing the tree.

\section{Methodology}
\label{sec:method}

DARTree is a training-free speculative decoding method that extends a pretrained causally corrected block-parallel drafter from a single chain to a speculative tree, without assuming a particular correction architecture. The central challenge is that branch scores become path-dependent after causal correction, making exact best-first construction inherently sequential. DARTree instead expands a fixed number of nodes in parallel at each depth, constructs a wider candidate supertree, and then prunes it into a compact tree for target-model verification. DARTree modifies only candidate-tree construction; the target-model tree-verification procedure is inherited unchanged from prior work. Figure~\ref{fig:method-overview} illustrates this workflow.

\subsection{Decoupling Causal Correction from Best-First Search}

Causal correction makes tree construction path-dependent: the children of a prefix cannot be scored until the correction head has processed the tokens along that prefix. A natural exact approach is to combine these corrected scores with the DDTree-style best-first construction~\cite{ringel2026accelerating}, maintaining a max-heap over candidate prefixes. At each step, the heap pops the highest-scoring prefix, the correction head scores its children under the corresponding token history, and these children are pushed back into the heap before the next prefix can be expanded. This interleaving preserves global best-first order under the corrected scores, but prevents correction from being batched across nodes.

DARTree relaxes the node-wise search order while preserving path-wise causal correction. It evaluates all branches at the same depth together, uses their corrected scores to retain the nodes for the next depth, and carries the selected correction states forward. Search and correction thus remain coupled across depths, but no heap decision separates two corrections within the same depth. The approximation lies only in replacing global, node-wise best-first expansion with a fixed-width, depth-wise schedule; every materialized branch is still scored by the pretrained correction head using its realized prefix.

\subsection{Depth-Wise Parallel Supertree Construction}

The block-parallel backbone is executed once to produce shared representations $\{H_d\}_{d=1}^{\gamma}$ for the entire draft block. To limit correction-head overhead, only its top-$K$ token predictions at each position are considered as candidate extensions. Starting from the root, DARTree batches the active branches at each depth and applies the pretrained correction module to their respective prefix states. It then ranks the candidate extensions by their corrected cumulative scores, retains the highest-scoring extensions under a fixed layer width, and carries the associated correction states to the next depth.

Repeating this depth-wise expansion yields a fixed-width candidate supertree. The fixed layer width gives every depth a regular batched workload: dependencies remain sequential across depths, whereas correction, scoring, selection, and state updates within each depth are tensorized. Because the lightweight correction head is evaluated in batch, increasing the layer width incurs little additional drafting latency, as shown in Figure~\ref{fig:dartree_budget_parallelism_latency}. This makes it practical to construct a supertree wider than the final verification tree. We introduce a verification budget $B$ to control the total number of nodes retained in the final tree passed to the target model. DARTree can therefore construct a wider supertree during drafting and prune it to $B$ only after all candidate scores are available.

\begin{algorithm}[t]
\captionsetup{labelfont=bf,textfont=bf}
\caption{DARTree Construction}
\label{alg:dartree}
\small
\begin{algorithmic}[1]
\REQUIRE Shared representations $\{H_d\}_{d=1}^{\gamma}$ and base logits $\{L_d^{\mathrm{base}}\}_{d=1}^{\gamma}$ from one block-parallel pass; correction head $g$; candidate-vocabulary size $K$; layer width $W$; node budget $B$; depth bonus $\beta\leq0$
\STATE Initialize the root layer $\mathcal S_0\gets\{\emptyset\}$, root correction state $r(\emptyset)\gets r_0$, and root score $s_\beta(\emptyset)\gets0$
\STATE Initialize the candidate supertree $\mathcal S\gets\emptyset$
\FOR{$d=1$ to $\gamma$}
    \STATE Select $\mathcal C_d\gets\operatorname{TopK}(L_d^{\mathrm{base}},K)$
    \STATE Correct all prefixes $u_{<d}\in\mathcal S_{d-1}$ over $\mathcal C_d$ in parallel:
    \STATE \hspace{\algorithmicindent}$\widetilde q_d(\mathcal C_d\mid u_{<d})\gets g_{\mathcal C_d}(H_d,r(u_{<d}))$
    \STATE Score each child $u_{1:d}=(u_{<d},u_d)$ with $u_d\in\mathcal C_d$ by
    \STATE \hspace{\algorithmicindent}$s_\beta(u_{1:d})\gets s_\beta(u_{<d})+\log\widetilde q_d(u_d\mid u_{<d})+\beta$
    \STATE Retain the global top-$W$ child prefixes as $\mathcal S_d$
    \STATE Update their states $r(u_{1:d})$ from $r(u_{<d})$ and $u_d$ in parallel, and set $\mathcal S\gets\mathcal S\cup\mathcal S_d$
\ENDFOR
\STATE Prune the complete supertree: $\mathcal T\gets\underset{u\in\mathcal S}{\operatorname{TopB}}\,s_\beta(u)$
\STATE \textbf{return} $\mathcal T$
\end{algorithmic}
\end{algorithm}

\subsection{Deferred Best-First Pruning}

Once the supertree is complete, every materialized prefix has a corrected score, so DARTree can recover best-first budget allocation without interleaving heap operations with correction. Let $u_{1:d}$ denote a depth-$d$ prefix and $\widetilde q_i(u_i)$ the corrected probability of its token at depth $i$, conditioned on the verified context and its preceding branch tokens. We score each prefix by
\begin{equation}
    s_\beta(u_{1:d})
    =
    \sum_{i=1}^{d}\log\widetilde q_i(u_i)+\beta d,
\label{eq:dartree-score}
\end{equation}
where $\beta$ is a depth bonus, with a negative value penalizing deeper prefixes.

\noindent\textbf{Lemma 1 (Heap--top-\(B\) equivalence).}
{\em Let $\mathcal S$ be a materialized prefix tree and $B$ the verification budget. If $\beta\leq0$ and ancestors are preferred under ties, best-first heap selection~\cite{ringel2026accelerating} and global top-\(B\) selection return the same tree:}
\begin{equation}
    \operatorname{BestFirst}_B(\mathcal S)
    =
    \underset{u\in\mathcal S}{\operatorname{TopB}}\,
    s_\beta(u).
\end{equation}
Extending a prefix adds a log-probability no greater than zero and a non-positive depth bonus, so no child can score above its parent. The globally highest-scoring $B$ nodes are therefore prefix-closed and are exactly those selected by a best-first max-heap sequentially. Hence, after the supertree has been scored, the heap can be replaced by a single top-\(B\) operation. The equivalence need not hold for a positive depth bonus, which may allow a descendant to outrank its ancestor. The resulting prefix tree is then verified by the target model in a single forward pass using a tree-attention mask. Algorithm~\ref{alg:dartree} summarizes the complete depth-wise construction and deferred pruning procedure. In the algorithm, each depth-$d$ node is identified by its root-to-node token prefix $u_{1:d}$; $\mathcal S_d$ contains only the retained nodes at depth $d$, whereas $\mathcal S$ accumulates the retained nodes across all depths.

\section{Experiments}
\label{sec:exp}
\begin{table*}[t]
  \centering
  \begingroup
  \fontsize{9}{10.5}\selectfont
  \setlength{\tabcolsep}{1mm}
  \renewcommand{\arraystretch}{1.03}
  \begin{tabular}{@{}ll*{8}{c@{\hspace{1mm}}c}@{}}
    \toprule
    \multirow{2}{*}{Model} & \multirow{2}{*}{Method}
      & \multicolumn{6}{c}{\textsc{Math}}
      & \multicolumn{4}{c}{\textsc{Code}}
      & \multicolumn{4}{c}{\textsc{Chat}}
      & \multicolumn{2}{c}{\textsc{Overall}} \\
    \cmidrule(lr){3-8}\cmidrule(lr){9-12}\cmidrule(lr){13-16}\cmidrule(lr){17-18}
      & & \multicolumn{2}{c}{GSM8K}
      & \multicolumn{2}{c}{MATH-500}
      & \multicolumn{2}{c}{AIME25}
      & \multicolumn{2}{c}{HumanEval}
      & \multicolumn{2}{c}{MBPP}
      & \multicolumn{2}{c}{MT-Bench}
      & \multicolumn{2}{c}{Alpaca}
      & \multicolumn{2}{c}{Avg.} \\
    \midrule
    \multicolumn{2}{c}{Temperature $=0$}
      & \makebox[0pt][c]{Speedup} & $\tau$ & \makebox[0pt][c]{Speedup} & $\tau$ & \makebox[0pt][c]{Speedup} & $\tau$
      & \makebox[0pt][c]{Speedup} & $\tau$ & \makebox[0pt][c]{Speedup} & $\tau$ & \makebox[0pt][c]{Speedup} & $\tau$
      & \makebox[0pt][c]{Speedup} & $\tau$ & \makebox[0pt][c]{Speedup} & $\tau$ \\
    \midrule
    \multirow{5}{*}{\shortstack{Qwen3\\4B}}
      & DFlash
      & 5.24$\times$ & 6.53 & 6.02$\times$ & 7.81 & 5.78$\times$ & 7.21
      & 5.13$\times$ & 6.59 & 4.79$\times$ & 6.08
      & 2.88$\times$ & 4.43 & 2.21$\times$ & 3.11 & 4.58$\times$ & 5.97 \\
      & DDTree
      & 6.62$\times$ & 8.47 & 7.26$\times$ & 9.71 & 6.73$\times$ & 8.99
      & 6.65$\times$ & 8.74 & 6.21$\times$ & 8.13
      & 3.98$\times$ & 6.01 & 3.11$\times$ & 4.39 & 5.79$\times$ & 7.78 \\
      & Domino
      & 7.37$\times$ & 10.14 & 6.71$\times$ & 9.05 & 5.72$\times$ & 7.38
      & 5.20$\times$ & 7.04 & 5.35$\times$ & 7.10
      & 3.16$\times$ & 5.22 & 2.68$\times$ & 3.87 & 5.17$\times$ & 7.11 \\
      & \mbox{DARTree (fixed)}
      & 9.06$\times$ & 12.74 & 8.25$\times$ & 11.64 & 7.20$\times$ & 10.05
      & 6.75$\times$ & 9.45 & 6.94$\times$ & 9.57
      & 4.10$\times$ & 6.84 & 3.52$\times$ & 5.23 & 6.55$\times$ & 9.36 \\
      & \mbox{DARTree (pruned)}
      & \textbf{9.73$\times$} & \textbf{12.97} & \textbf{8.49$\times$} & \textbf{12.05} & \textbf{7.45$\times$} & \textbf{10.54}
      & \textbf{7.36$\times$} & \textbf{9.94} & \textbf{7.34$\times$} & \textbf{10.08}
      & \textbf{4.62$\times$} & \textbf{7.34} & \textbf{3.93$\times$} & \textbf{5.72} & \textbf{6.99$\times$} & \textbf{9.81} \\
    \cmidrule(lr){1-18}
    \multirow{5}{*}{\shortstack{Qwen3\\8B}}
      & DFlash
      & 4.99$\times$ & 6.55 & 5.93$\times$ & 7.93 & 5.59$\times$ & 7.25
      & 5.02$\times$ & 6.58 & 4.51$\times$ & 5.99
      & 2.68$\times$ & 4.25 & 2.18$\times$ & 3.12 & 4.41$\times$ & 5.95 \\
      & DDTree
      & 5.94$\times$ & 8.44 & 6.84$\times$ & 9.76 & 6.17$\times$ & 8.81
      & 6.08$\times$ & 8.61 & 5.55$\times$ & 7.95
      & 3.52$\times$ & 5.80 & 2.92$\times$ & 4.42 & 5.29$\times$ & 7.68 \\
      & Domino
      & 7.23$\times$ & 10.00 & 6.61$\times$ & 9.36 & 5.40$\times$ & 7.48
      & 5.34$\times$ & 7.34 & 5.06$\times$ & 7.05
      & 3.04$\times$ & 5.18 & 2.50$\times$ & 3.80 & 5.03$\times$ & 7.17 \\
      & \mbox{DARTree (fixed)}
      & 8.47$\times$ & 12.57 & 7.89$\times$ & 11.86 & 6.91$\times$ & 10.20
      & 6.66$\times$ & 9.81 & 6.33$\times$ & 9.56
      & 3.82$\times$ & 6.85 & 3.25$\times$ & 5.25 & 6.19$\times$ & 9.44 \\
      & \mbox{DARTree (pruned)}
      & \textbf{8.64$\times$} & \textbf{12.93} & \textbf{8.14$\times$} & \textbf{12.25} & \textbf{7.14$\times$} & \textbf{10.55}
      & \textbf{6.98$\times$} & \textbf{10.37} & \textbf{6.71$\times$} & \textbf{10.09}
      & \textbf{4.24$\times$} & \textbf{7.43} & \textbf{3.57$\times$} & \textbf{5.74} & \textbf{6.49$\times$} & \textbf{9.91} \\
    \midrule
    \multicolumn{2}{c}{Temperature $=1$}
      & \makebox[0pt][c]{Speedup} & $\tau$ & \makebox[0pt][c]{Speedup} & $\tau$ & \makebox[0pt][c]{Speedup} & $\tau$
      & \makebox[0pt][c]{Speedup} & $\tau$ & \makebox[0pt][c]{Speedup} & $\tau$ & \makebox[0pt][c]{Speedup} & $\tau$
      & \makebox[0pt][c]{Speedup} & $\tau$ & \makebox[0pt][c]{Speedup} & $\tau$ \\
    \midrule
    \multirow{5}{*}{\shortstack{Qwen3\\4B}}
      & DFlash
      & 4.60$\times$ & 5.94 & 5.14$\times$ & 6.67 & 3.81$\times$ & 4.91
      & 4.79$\times$ & 6.01 & 4.35$\times$ & 5.60
      & 2.59$\times$ & 4.00 & 2.15$\times$ & 3.00 & 3.92$\times$ & 5.16 \\
      & DDTree
      & 5.96$\times$ & 7.92 & 6.49$\times$ & 8.63 & 4.92$\times$ & 6.80
      & 6.19$\times$ & 8.03 & 5.84$\times$ & 7.71
      & 3.60$\times$ & 5.50 & 2.95$\times$ & 4.14 & 5.14$\times$ & 6.96 \\
      & Domino
      & 6.26$\times$ & 8.32 & 5.20$\times$ & 7.14 & 3.43$\times$ & 4.70
      & 4.71$\times$ & 6.20 & 4.99$\times$ & 6.26
      & 2.69$\times$ & 4.49 & 2.36$\times$ & 3.33 & 4.23$\times$ & 5.78 \\
      & \mbox{DARTree (fixed)}
      & 8.03$\times$ & 11.17 & 6.63$\times$ & 9.60 & 4.50$\times$ & 6.56
      & 6.30$\times$ & 8.65 & 6.39$\times$ & 8.85
      & 3.73$\times$ & 6.20 & 3.19$\times$ & 4.72 & 5.54$\times$ & 7.96 \\
      & \mbox{DARTree (pruned)}
      & \textbf{8.26$\times$} & \textbf{11.63} & \textbf{6.92$\times$} & \textbf{10.10} & \textbf{5.01$\times$} & \textbf{7.37}
      & \textbf{6.76$\times$} & \textbf{9.11} & \textbf{6.88$\times$} & \textbf{9.33}
      & \textbf{4.09$\times$} & \textbf{6.61} & \textbf{3.61$\times$} & \textbf{5.23} & \textbf{5.93$\times$} & \textbf{8.48} \\
    \cmidrule(lr){1-18}
    \multirow{5}{*}{\shortstack{Qwen3\\8B}}
      & DFlash
      & 4.44$\times$ & 5.92 & 4.60$\times$ & 6.34 & 3.54$\times$ & 4.81
      & 4.22$\times$ & 5.54 & 3.97$\times$ & 5.29
      & 2.41$\times$ & 3.78 & 2.08$\times$ & 3.01 & 3.61$\times$ & 4.96 \\
      & DDTree
      & 5.47$\times$ & 7.82 & 5.76$\times$ & 8.43 & \textbf{4.56$\times$} & 6.66
      & 5.38$\times$ & 7.64 & 5.01$\times$ & 7.22
      & 3.19$\times$ & 5.21 & 2.80$\times$ & 4.24 & 4.60$\times$ & 6.75 \\
      & Domino
      & 5.79$\times$ & 8.14 & 4.85$\times$ & 7.10 & 3.40$\times$ & 4.78
      & 4.22$\times$ & 5.87 & 4.14$\times$ & 5.84
      & 2.49$\times$ & 4.28 & 2.13$\times$ & 3.20 & 3.86$\times$ & 5.60 \\
      & \mbox{DARTree (fixed)}
      & 7.31$\times$ & 11.04 & 6.07$\times$ & 9.45 & 4.43$\times$ & 6.76
      & 5.45$\times$ & 8.06 & 5.45$\times$ & 8.21
      & 3.29$\times$ & 5.84 & 2.96$\times$ & 4.73 & 4.99$\times$ & 7.73 \\
      & \mbox{DARTree (pruned)}
      & \textbf{7.34$\times$} & \textbf{11.12} & \textbf{6.41$\times$} & \textbf{9.88} & 4.50$\times$ & \textbf{6.98}
      & \textbf{5.65$\times$} & \textbf{8.42} & \textbf{5.68$\times$} & \textbf{8.63}
      & \textbf{3.61$\times$} & \textbf{6.29} & \textbf{3.22$\times$} & \textbf{5.16} & \textbf{5.20$\times$} & \textbf{8.07} \\
    \bottomrule
  \end{tabular}
  \endgroup
  \caption{\textbf{DARTree results on Qwen3-4B and Qwen3-8B.}
  We report decoding speedup over locally measured vanilla autoregressive decoding and average acceptance length ($\tau$, the number of tokens accepted per verification round).
  The \textbf{best} result in each model and temperature block is highlighted in bold.}
  \label{tab:dartree-main-results}
\end{table*}

\subsection{Experimental Setup}

\noindent\textbf{Models and Benchmarks.}
Following prior work~\cite{chen2026dflash,huang2026domino}, we evaluate DARTree on Qwen3-4B and Qwen3-8B~\cite{yang2025qwen3}. 
We report the average acceptance length $\tau$, defined as the number of output tokens advanced per draft--verify round, including the bonus token, together with speedup over local AR decoding. Let $T_{\mathrm{draft}}$ include draft generation and tree construction, and let $T_{\mathrm{verify}}$ denote target verification latency. The amortized per-token latency and speedup are
\begin{equation}
    L_{\mathrm{spec}}
    = \frac{T_{\mathrm{draft}}+T_{\mathrm{verify}}}{\tau},
    \qquad
    \eta = \frac{L_{\mathrm{AR}}}{L_{\mathrm{spec}}}.
\end{equation}
Consistent with the evaluation protocols used by existing methods, we first compute internal decoding latency per sample and then macro-average across samples.
Our benchmarks span three domains: \textbf{Math}: GSM8K~\cite{cobbe2021training}, MATH-500~\cite{lightman2023lets}, and AIME25~\cite{aime25}; \textbf{Code}: HumanEval~\cite{chen2021evaluating}, MBPP~\cite{austin2021program}; and \textbf{Chat}: MT-Bench~\cite{zheng2023judging} and Alpaca~\cite{taori2023alpaca}.

\noindent\textbf{Baselines and Implementation.}
We evaluate DARTree with vanilla AR decoding and representative diffusion-based drafters: DFlash~\cite{chen2026dflash}, which predicts a 16-token block in parallel; DDTree~\cite{ringel2026accelerating}, which builds a best-first tree from DFlash's marginal predictions; and Domino~\cite{huang2026domino}, which jointly trains a DFlash backbone with an autoregressive correction head.
By default, DARTree uses the released Domino correction head without additional training. We evaluate two variants: \textbf{DARTree (fixed)} assigns the same number of nodes to each depth and directly uses the resulting tree for verification, whereas \textbf{DARTree (pruned)} constructs a wider supertree and subsequently applies deferred pruning to meet the target-verification budget. Both DARTree variants use $K=64$ candidate tokens per position. Unless otherwise specified, DDTree and both DARTree variants use the same $B=64$ target-verification budget. DARTree (fixed) distributes this budget uniformly across 16 draft depths, retaining 4 nodes at each depth.  For DARTree (pruned), we use a supertree width of 12 and a pruning depth bonus of $\beta=-0.2$. We additionally examine the effects of the layer width and verification budget through hyperparameter sweeps.

\noindent\textbf{Evaluation Details.}
We evaluate greedy decoding at $T=0$ and, following DFlash~\cite{chen2026dflash} and DDTree~\cite{ringel2026accelerating}, sample from the target model at $T=1$: verification accepts the longest sampled prefix in the draft tree and emits the target sample at the first mismatch. We use batch size one and generate at most 2048 tokens per prompt. All methods and baselines are measured on one NVIDIA RTX 6000 Ada Generation GPU.

\subsection{Main Results}

Table~\ref{tab:dartree-main-results} shows that our DARTree (fixed) already outperforms both DDTree and Domino on the overall averages in all four model--temperature configurations. It improves $\tau$ by up to 22.9\% over DDTree and 37.9\% over Domino, with corresponding speedup gains of up to 17.0\% and 30.8\%. This demonstrates that extending causal correction from a single chain to multiple depth-wise branches provides substantial gains even without adaptive tree pruning. Furthermore, it is evident that \textit{DARTree (pruned) achieves the highest overall average $\tau$ and speedup in all four configurations}, outperforming DDTree and Domino by up to 28.9\% and 46.8\% in $\tau$, and by up to 22.7\% and 40.1\% in speedup, respectively. The only exception for individual benchmarks is AIME25 with Qwen3-8B at $T=1$, where DDTree is marginally faster (4.56$\times$ versus 4.50$\times$), while DARTree still attains a higher acceptance length. Compared with the fixed variant, deferred pruning further improves $\tau$ by up to 6.5\% and speedup by up to 7.1\%, showing the benefit of selecting the verification tree from a wider set of corrected branches. The improvements at $T=1$ further show that DARTree's gains extend beyond greedy decoding to stochastic sampling.
On GSM8K with Qwen3-4B at $T=0$, DARTree accepts 12.97 tokens per verification round and reaches a 9.73$\times$ speedup. Its acceptance length is 98.6\% higher than DFlash and 27.9\% higher than Domino in the same setting.

\subsection{Ablation Study}

We isolate two aspects of DARTree: its depth-wise construction schedule and its compatibility with different causal-correction heads. The construction ablations use Qwen3-4B at $T=0$ and the same 64-node target-verification budget.

\paragraph{Tree-construction Strategy.}
In Table~\ref{tab:construction-ablation}, we compare DARTree with two natural alternatives. Sequential Correction w. Heap interleaves correction-head inference with best-first heap updates. It obtains acceptance lengths close to DARTree, confirming the quality of fully conditional node-wise search, but requires roughly 70\,ms per round, more than twice DARTree's latency on all three datasets. Domino-chain + DDTree first applies causal correction along a single draft chain and then uses the resulting chain-conditioned distributions to construct a tree with DDTree. Although efficient, alternative branches are not corrected according to their own token histories. By applying path-specific correction to multiple branches in parallel, DARTree improves $\tau$ over Domino-chain + DDTree by up to 16.4\%, while avoiding the sequential correction and heap-update overhead of the first alternative. It therefore achieves the highest speedup on all three datasets.

\begin{table}[!t]
  \centering
  \begingroup
  \small
  \setlength{\tabcolsep}{1.8pt}
  \begin{tabular}{lccc}
    \toprule
    Method & $\tau$ & Round Time (ms) & Speedup \\
    \midrule
    \multicolumn{4}{l}{\textbf{GSM8K}} \\
    Seq. Correction w. Heap       & 12.840 & 62.703{\scriptsize $\pm 0.695$} & 4.38{\scriptsize $\pm 0.05$}$\times$ \\
    Domino-chain + DDTree          & 11.272 & 28.285{\scriptsize $\pm 0.207$} & 8.48{\scriptsize $\pm 0.06$}$\times$ \\
    DARTree              & \textbf{12.967} & 28.494{\scriptsize $\pm 0.606$} & \textbf{9.77}{\scriptsize $\mathbf{\pm 0.21}$}$\times$ \\
    \addlinespace[2pt]
    \midrule
    \multicolumn{4}{l}{\textbf{HumanEval}} \\
    Seq. Correction w. Heap       & 9.827 & 64.267{\scriptsize $\pm 0.876$} & 3.25{\scriptsize $\pm 0.04$}$\times$ \\
    Domino-chain + DDTree          & 8.539 & 28.643{\scriptsize $\pm 0.276$} & 6.31{\scriptsize $\pm 0.06$}$\times$ \\
    DARTree              & \textbf{9.938} & 28.805{\scriptsize $\pm 0.409$} & \textbf{7.36}{\scriptsize $\mathbf{\pm 0.11}$}$\times$ \\
    \addlinespace[2pt]
    \midrule
    \multicolumn{4}{l}{\textbf{MT-Bench}} \\
    Seq. Correction w. Heap       & \textbf{7.369} & 65.828{\scriptsize $\pm 0.702$} & 2.01{\scriptsize $\pm 0.02$}$\times$ \\
    Domino-chain + DDTree          & 6.520 & 29.502{\scriptsize $\pm 0.180$} & 3.98{\scriptsize $\pm 0.03$}$\times$ \\
    DARTree              & 7.343 & 29.098{\scriptsize $\pm 0.437$} & \textbf{4.56}{\scriptsize $\mathbf{\pm 0.07}$}$\times$ \\
    \bottomrule
  \end{tabular}
  \endgroup
  \caption{Ablation of tree construction strategies on Qwen3-4B at temperature 0. Round time and speedup are computed per sample and macro-averaged over each benchmark}
  \label{tab:construction-ablation}
\end{table}

\begin{table}[!t]
  \centering
  \begingroup
  \small
  \setlength{\tabcolsep}{4.5pt}
  \begin{tabular}{lcc@{\hspace{8pt}}cc}
    \toprule
    Method
      & \multicolumn{2}{c}{Temperature $=0$}
      & \multicolumn{2}{c}{Temperature $=1$} \\
    \cmidrule(lr){2-3}\cmidrule(lr){4-5}
      & $\tau$ & Speedup & $\tau$ & Speedup \\
    \midrule
    \multicolumn{5}{l}{\textbf{GSM8K}} \\
    DSpark-Markov
      & 6.607 & 5.54$\times$ & 5.732 & 4.66$\times$ \\
    + DARTree (pruned)
      & \textbf{7.571} & \textbf{5.61$\times$}
      & \textbf{7.319} & \textbf{5.26$\times$} \\
    \addlinespace[2pt]
    \midrule
    \multicolumn{5}{l}{\textbf{HumanEval}} \\
    DSpark-Markov
      & 5.728 & 4.78$\times$ & 5.040 & 3.97$\times$ \\
    + DARTree (pruned)
      & \textbf{7.112} & \textbf{5.37$\times$}
      & \textbf{6.820} & \textbf{4.94$\times$} \\
    \addlinespace[2pt]
    \midrule
    \multicolumn{5}{l}{\textbf{MT-Bench}} \\
    DSpark-Markov
      & 4.437 & 3.31$\times$ & 3.881 & 2.71$\times$ \\
    + DARTree (pruned)
      & \textbf{5.825} & \textbf{3.97$\times$}
      & \textbf{5.455} & \textbf{3.64$\times$} \\
    \bottomrule
  \end{tabular}
  \endgroup
  \caption{Ablation using the released DSpark 
with Markov correction head on Qwen3-4B.}
  \label{tab:correction-head-ablation}
\end{table}

\paragraph{Correction-head Transfer.}
To evaluate the extensibility of DARTree across different causally corrected diffusion drafters, we additionally apply it to the Markov-head variant of the recently released diffusion drafter DSpark~\cite{cheng2026dspark}. As shown in Table~\ref{tab:correction-head-ablation}, DARTree improves both metrics in all six dataset--temperature settings: acceptance length increases by 14.6--40.6\%, while speedup improves by up to 34.3\%. These results support the view that DARTree is a general tree-construction method for causally corrected parallel drafters rather than a Domino-specific modification.

\begin{figure}[!t]
  \centering
  \includegraphics[width=\columnwidth]{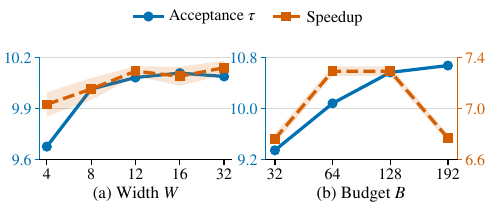}
  \caption{Effect of (a) layer width $W$ with fixed $B=64$ and (b) verification budget $B$ with fixed $W=12$, averaged over three benchmarks. Shading indicates propagated standard deviation across four runs.}
  \label{fig:sensitivity-sweeps}
\end{figure}

\paragraph{Influence of Hyperparameters.}
We study the two important hyperparameters of DARTree on Qwen3-4B at $T=0$, averaging equally over GSM8K, HumanEval, and MT-Bench. Figure~\ref{fig:sensitivity-sweeps}(a) examines the effect of layer width while fixing $B=64$. Increasing the width from 4 to 12 noticeably improves acceptance by preserving more branch alternatives, whereas wider layers provide little additional benefit.
Figure~\ref{fig:sensitivity-sweeps}(b) fixes $W=12$ and varies the target-verification budget. Larger budgets monotonically improve acceptance over most of the tested range, but speedup peaks around $B=64$--128 and falls at $B=192$. We therefore use $B=64$ in the main experiments as a balanced operating point.

\begin{figure}[!t]
  \centering
  \includegraphics[width=\columnwidth]{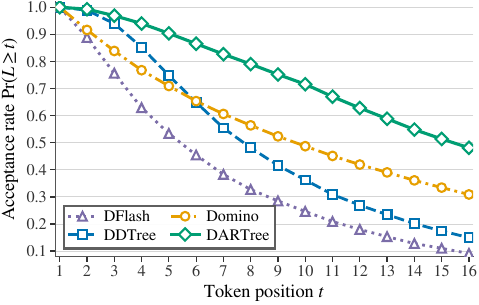}
  \caption{Position-wise acceptance rates on GSM8K. Here, $L$ denotes the acceptance length of a decoding round, and $\Pr(L\geq t)$ is the probability that a decoding round advances at least $t$ output tokens. DARTree is much more likely to reach later positions.}
  \label{fig:acceptance-distribution}
\end{figure}

\paragraph{Acceptance Patterns.}
Figure~\ref{fig:acceptance-distribution} examines where the acceptance improvement arises. DARTree maintains a visibly higher probability of reaching later positions: on GSM8K, DARTree accepts almost the entire draft block in nearly half of the decoding rounds. DDTree provides stronger acceptance at early positions, whereas Domino maintains a higher probability of reaching later positions; DARTree combines both advantages, preserving strong early acceptance while substantially extending the accepted continuation.

\section{Conclusion}

We presented DARTree, a training-free speculative decoding method that extends causally corrected block-parallel drafting from a single chain to a tree. DARTree carries path-specific correction across multiple branches through depth-wise batched expansion, and performs deferred best-first pruning with a single global top-$B$ selection after the candidate supertree has been fully scored. This design avoids the node-wise coupling of correction-head inference and sequential heap updates while leaving the standard target-model tree-verification procedure unchanged. Across all evaluated benchmarks, DARTree achieves the highest overall average acceptance length and speedup in every model--temperature configuration, with consistent benefits under both greedy decoding and stochastic sampling.


\newpage
\bibliography{main}

\clearpage
\setcounter{secnumdepth}{2}

\appendix
\section*{\Large{Appendix}}

\subsection*{A. Limitations}
Despite its consistent performance across a wide range of evaluations, DARTree has several limitations. First, DARTree relies on a pretrained diffusion drafter equipped with a causal correction head. Consequently, its training-free property does not directly apply to naive diffusion drafters, which require additional training to incorporate such a head. This requirement also restricts our evaluation to compatible models publicly available.
Moreover, as a speculative decoding method, DARTree does not reduce the total number of FLOPs; instead, it uses additional computation to reduce inference latency. Similar to other tree-based speculative decoding approaches, DARTree incurs even greater computational overhead than non-tree-based methods because it verifies a large tree of candidate tokens. Therefore, it may not be suitable for all deployment environments. We discuss the scenarios in which DARTree is most effective in the following section.

\subsection*{B. When Should We Use DARTree?}
Like other tree-based speculative decoding methods, DARTree is primarily designed for low-concurrency serving, where decoding is highly memory-bandwidth-bound and the computational resources remain underutilized. In this regime, additional computation for drafting and tree verification can be traded for lower decoding latency and higher token throughput. As concurrency increases, batched decoding improves hardware utilization and the workload becomes increasingly compute-bound, making the overhead of verifying a large candidate tree more costly. Table~\ref{tab:concurrency} evaluates this trade-off across different concurrency levels.

\begin{table*}[t]
    \centering
    \small
    \setlength{\tabcolsep}{4pt}
    \begin{tabular}{lccccc}
        \toprule
        Method & \(C=2\) & \(C=4\) & \(C=8\) & \(C=16\) & \(C=32\) \\
        \midrule
        \multicolumn{6}{l}{\textbf{GSM8K}} \\
        AR
        & 159.5 / 1.00\(\times\)
        & 311.9 / 1.00\(\times\)
        & 582.0 / 1.00\(\times\)
        & 1046.8 / 1.00\(\times\)
        & 1645.8 / 1.00\(\times\) \\
        EAGLE-3 (\(B=64\))
        & 332.6 / 2.08\(\times\)
        & 462.9 / 1.48\(\times\)
        & 535.4 / 0.92\(\times\)
        & 576.4 / 0.55\(\times\)
        & 593.9 / 0.36\(\times\) \\
        DFlash
        & 653.7 / 4.10\(\times\)
        & 1199.4 / 3.85\(\times\)
        & 1886.3 / 3.24\(\times\)
        & 2497.5 / 2.39\(\times\)
        & 2761.7 / 1.68\(\times\) \\
        Domino
        & 923.4 / 5.79\(\times\)
        & 1631.5 / 5.23\(\times\)
        & 2292.2 / 3.94\(\times\)
        & 2801.3 / 2.68\(\times\)
        & 3387.8 / 2.06\(\times\) \\
        DDTree (\(B=64\))
        & 709.7 / 4.45\(\times\)
        & 859.5 / 2.76\(\times\)
        & 1001.3 / 1.72\(\times\)
        & 1173.7 / 1.12\(\times\)
        & 1264.8 / 0.77\(\times\) \\
        DDTree (adaptive \(B\))
        & 709.7 / 4.45\(\times\)
        & 1314.5 / 4.21\(\times\)
        & 2197.3 / 3.78\(\times\)
        & 2614.3 / 2.50\(\times\)
        & 2873.9 / 1.75\(\times\) \\
        DARTree (\(B=64,W=12\))
        & 987.5 / 6.19\(\times\)
        & 1212.4 / 3.89\(\times\)
        & 1379.9 / 2.37\(\times\)
        & 1638.5 / 1.57\(\times\)
        & 1749.2 / 1.06\(\times\) \\
        DARTree (adaptive \(B/W\))
        & \textbf{987.5 / 6.19\(\times\)}
        & \textbf{1766.2 / 5.66\(\times\)}
        & \textbf{2806.6 / 4.82\(\times\)}
        & \textbf{3395.0 / 3.24\(\times\)}
        & \textbf{3701.5 / 2.25\(\times\)} \\
        \midrule
        \multicolumn{6}{l}{\textbf{HumanEval}} \\
        AR
        & 159.1 / 1.00\(\times\)
        & 309.6 / 1.00\(\times\)
        & 571.3 / 1.00\(\times\)
        & 977.1 / 1.00\(\times\)
        & 1301.6 / 1.00\(\times\) \\
        EAGLE-3 (\(B=64\))
        & 304.9 / 1.92\(\times\)
        & 422.9 / 1.37\(\times\)
        & 491.2 / 0.86\(\times\)
        & 516.0 / 0.53\(\times\)
        & 540.4 / 0.42\(\times\) \\
        DFlash
        & 672.3 / 4.23\(\times\)
        & 1228.2 / 3.97\(\times\)
        & 2045.4 / 3.58\(\times\)
        & 2417.4 / 2.47\(\times\)
        & 2605.5 / 2.00\(\times\) \\
        Domino
        & 677.1 / 4.26\(\times\)
        & 1229.8 / 3.97\(\times\)
        & 1977.8 / 3.46\(\times\)
        & 2278.3 / 2.33\(\times\)
        & 2471.3 / 1.90\(\times\) \\
        DDTree (\(B=64\))
        & 729.9 / 4.59\(\times\)
        & 882.4 / 2.85\(\times\)
        & 1016.4 / 1.78\(\times\)
        & 1201.5 / 1.23\(\times\)
        & 1276.2 / 0.98\(\times\) \\
        DDTree (adaptive \(B\))
        & 729.9 / 4.59\(\times\)
        & 1327.1 / 4.29\(\times\)
        & 2116.2 / 3.70\(\times\)
        & 2557.1 / 2.62\(\times\)
        & 2692.6 / 2.07\(\times\) \\
        DARTree (\(B=64,W=12\))
        & 773.0 / 4.86\(\times\)
        & 938.4 / 3.03\(\times\)
        & 1101.1 / 1.93\(\times\)
        & 1304.9 / 1.34\(\times\)
        & 1391.4 / 1.07\(\times\) \\
        DARTree (adaptive \(B/W\))
        & \textbf{773.0 / 4.86\(\times\)}
        & \textbf{1358.9 / 4.39\(\times\)}
        & \textbf{2192.8 / 3.84\(\times\)}
        & \textbf{2624.3 / 2.69\(\times\)}
        & \textbf{2806.9 / 2.16\(\times\)} \\
        \bottomrule
    \end{tabular}
    \caption{\textbf{Concurrent-serving results on Qwen3-4B at \(T=0\).}
    All methods are evaluated using SGLang in BF16 on a single NVIDIA RTX 6000 Ada Generation. Throughput is measured as aggregate completion-token TPS using end-to-end request wall-clock time, including prompt prefill. Each cell reports TPS / speedup over AR at the same concurrency. For the adaptive DDTree and DARTree rows, \(B\) is set to \(64,32,16,16,16\) at \(C=2,4,8,16,32\), respectively; DARTree additionally uses \(W=12,8,3,3,3\). The same schedule is used for both datasets.}
    \label{tab:concurrency}
\end{table*}

The results show that DARTree delivers its largest gains at low concurrency. As concurrency increases, the benefit of a fixed large tree becomes smaller, while adjusting \(B\) and \(W\) to the serving load can maintain strong performance. These results highlight the importance of selecting an appropriate tree configuration for the deployment setting.

In summary, the recommended scenarios for DARTree are:
\begin{itemize}
    \item Local, personal, or small-scale model serving with one or a few concurrent requests, where per-request token throughput is the primary concern.
    \item Latency-critical or high-priority requests that are processed with small batch sizes.
    \item Other deployments with underutilized compute resources, where the additional cost of tree-based verification can be absorbed efficiently.
\end{itemize}

\subsection*{C. Proof of Lemma 1}

\noindent\textit{Proof.}
Assume that both procedures use the tie-breaking convention in Lemma~1.
For any node \(u=u_{1:d}\) with parent \(p(u)\),
\[
    s_\beta(u)-s_\beta(p(u))
    =
    \log \widetilde q_d(u_d\mid p(u))+\beta
    \leq 0,
\]
since \(\widetilde q_d(u_d\mid p(u))\leq1\) and \(\beta\leq0\).
Scores are therefore non-increasing along every root-to-leaf path and, with
ancestors preferred under ties, the global top-\(B\) set is prefix-closed.
Moreover, any unexposed node has a frontier ancestor in the heap with at
least the same score. Thus each heap pop returns the next node in the global
ordering, and the first \(B\) popped nodes are exactly the global top-\(B\)
nodes.

\subsection*{D. Additional Experiments}

\paragraph{Ablations on Depth Bonus and Candidate Sizes.}

\begin{figure}[!t]
    \centering
    \includegraphics[width=\columnwidth]{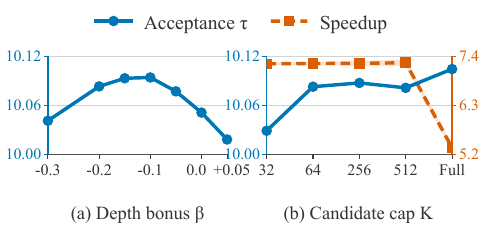}
    \caption{Effect of (a) the depth bonus \(\beta\) and (b) the
    candidate-vocabulary cap \(K\) on Qwen3-4B, averaged equally
    over GSM8K, HumanEval, and MT-Bench. ``Full''
    denotes the full vocabulary for AR-head correction scoring.}
    \label{fig:additional-hyperparameter-sensitivity}
\end{figure}

We further examine the depth bonus \(\beta\) and candidate-vocabulary cap
\(K\) on Qwen3-4B at \(T=0\). As shown in
Figure~\ref{fig:additional-hyperparameter-sensitivity}, the highest average
acceptance length across the three datasets is achieved with negative depth
bonuses between \(-0.2\) and \(-0.1\), while performance remains stable for
nearby values. Varying \(K\) from 32 to 512 has little effect on acceptance or
speedup. However, using the full vocabulary for tree expansion
substantially reduces speedup due to the increased latency of the AR head.
These results support \(\beta=-0.2\) and \(K=64\) as efficient default
settings.

\paragraph{Distribution of Verification-Tree Shapes.}

\begin{figure*}[!t]
    \centering
    \includegraphics[width=0.96\textwidth]{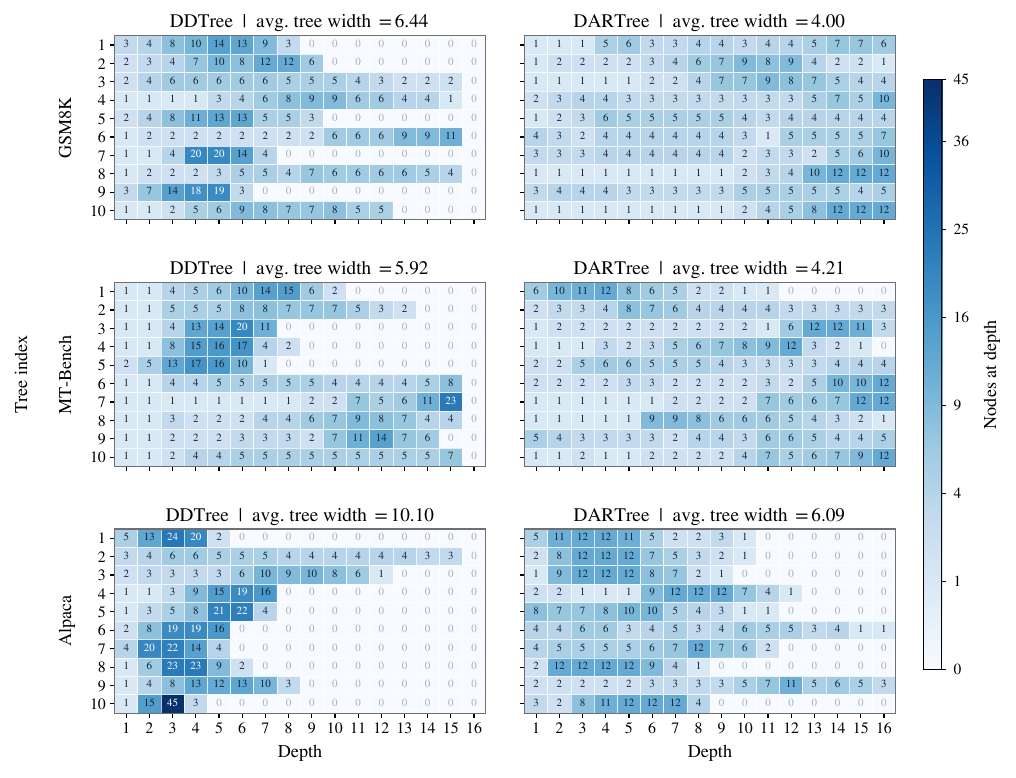}
    \caption{\textbf{Node-depth distributions of verification trees.}
    Each panel contains 10 example trees constructed with budget \(B=64\). Each row denotes an individual tree, and the number in each cell is the number of nodes allocated
    to that depth. }
    \label{fig:tree-node-depth-distributions}
\end{figure*}

Figure~\ref{fig:tree-node-depth-distributions} compares how DDTree and
DARTree allocate the verification budget across depths. Across all
three benchmarks, DARTree forms narrower trees and allocates more candidates to deeper levels, whereas DDTree more often concentrates its budget in a few
wide, shallow levels. 

\paragraph{Visualization of Example Trees.}

\begin{figure*}[p]
    \centering
    \includegraphics[page=1,width=\textwidth,height=0.82\textheight,
    keepaspectratio,trim=20 70 20 70,clip]
    {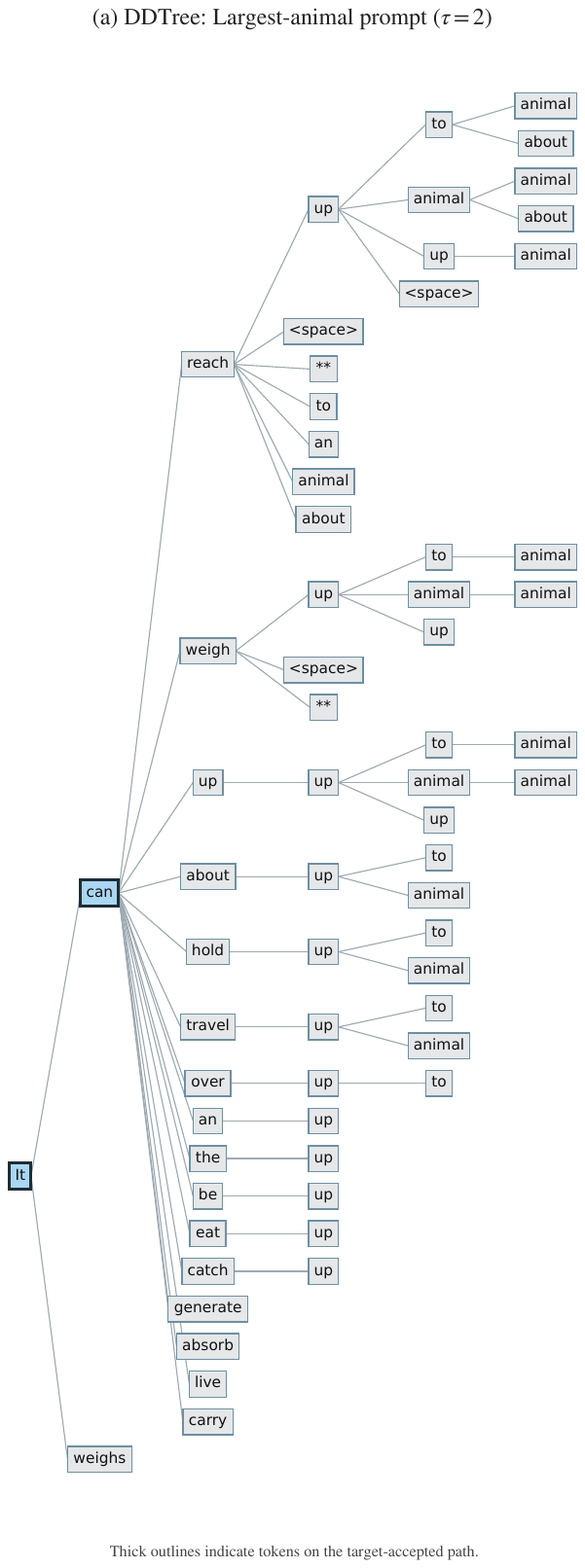}
    \caption{\textbf{DDTree verification tree for the largest-animal prompt.}
    The prompt is
    ``\texttt{Can you tell me what is the largest animal in the world?}''
    The tree is constructed after the complete prefix
    ``\texttt{The largest animal in the world is the blue whale (Balaenoptera
    musculus). Key facts about the blue whale: Size:}''. The target model
    accepts the two-token continuation
    \texttt{It can} (\(\tau=2\)). Blue nodes with thick outlines mark the
    accepted path; gray nodes with thin outlines mark the remaining candidates. DDTree allocates many
    nodes to a few depths, producing unusually wide layers.}
    \label{fig:ddtree-largest-animal-tree}
\end{figure*}

\begin{figure*}[p]
    \centering
    \includegraphics[page=2,width=\textwidth,height=0.82\textheight,
    keepaspectratio,trim=20 70 20 70,clip]
    {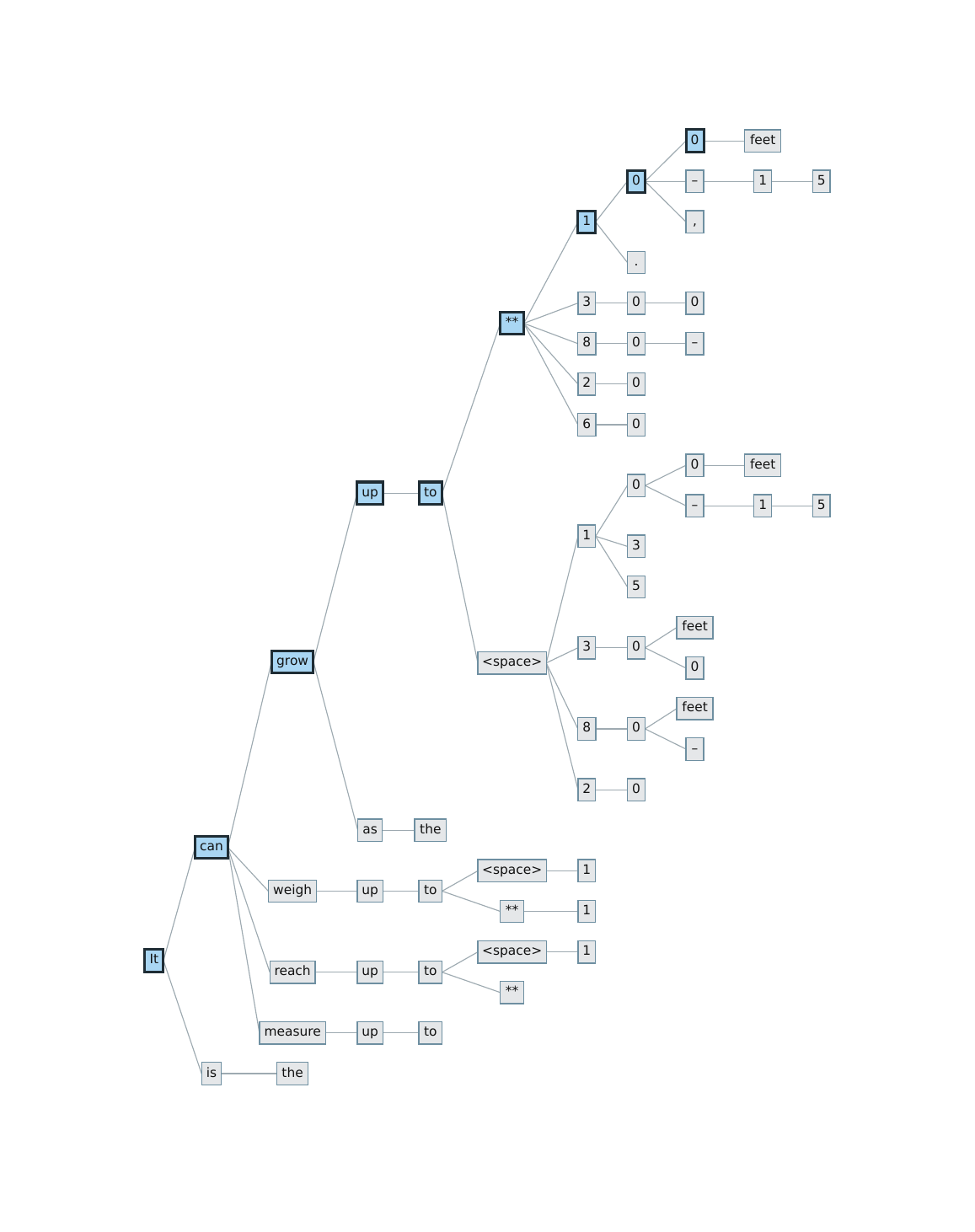}
    \caption{\textbf{DARTree verification tree for the largest-animal prompt.}
    The prompt is
    ``\texttt{Can you tell me what is the largest animal in the world?}''
    The tree is constructed after the complete prefix
    ``\texttt{The largest animal in the world is the blue whale (Balaenoptera
    musculus). Key facts about the blue whale: Size:}''. The target model
    accepts the nine-token continuation
    \texttt{It can grow up to **100} (\(\tau=9\)). Blue nodes with thick
    outlines mark the accepted path; gray nodes with thin outlines mark the
    remaining candidates.}
    \label{fig:dartree-largest-animal-tree}
\end{figure*}

\begin{figure*}[p]
    \centering
    \includegraphics[page=3,width=\textwidth,height=0.82\textheight,
    keepaspectratio,trim=20 70 20 70,clip]
    {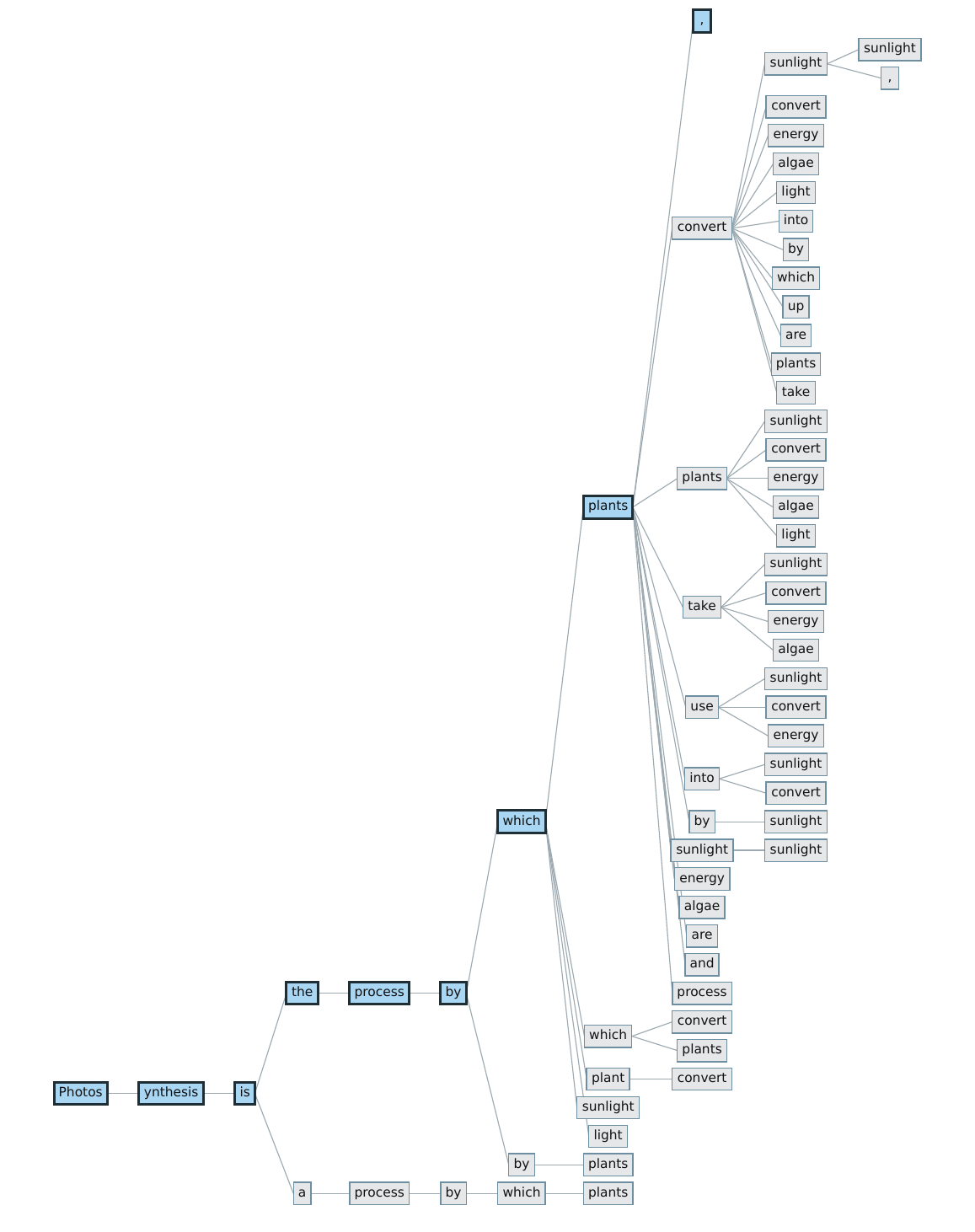}
    \caption{\textbf{DDTree verification tree for the photosynthesis prompt.}
    The prompt is
    ``\texttt{What is photosynthesis? Answer in exactly two short sentences.}''
    The tree begins at the first generated token. The target model accepts the nine-token
    continuation \texttt{Photosynthesis is the process by which plants,}
    (\(\tau=9\)). Blue nodes with thick outlines mark the accepted path; gray
    nodes with thin outlines mark the remaining candidates. DDTree allocates many nodes to a
    few depths, producing unusually wide layers.}
    \label{fig:ddtree-photosynthesis-tree}
\end{figure*}

\begin{figure*}[p]
    \centering
    \includegraphics[page=4,width=\textwidth,height=0.82\textheight,
    keepaspectratio,trim=20 70 20 70,clip]
    {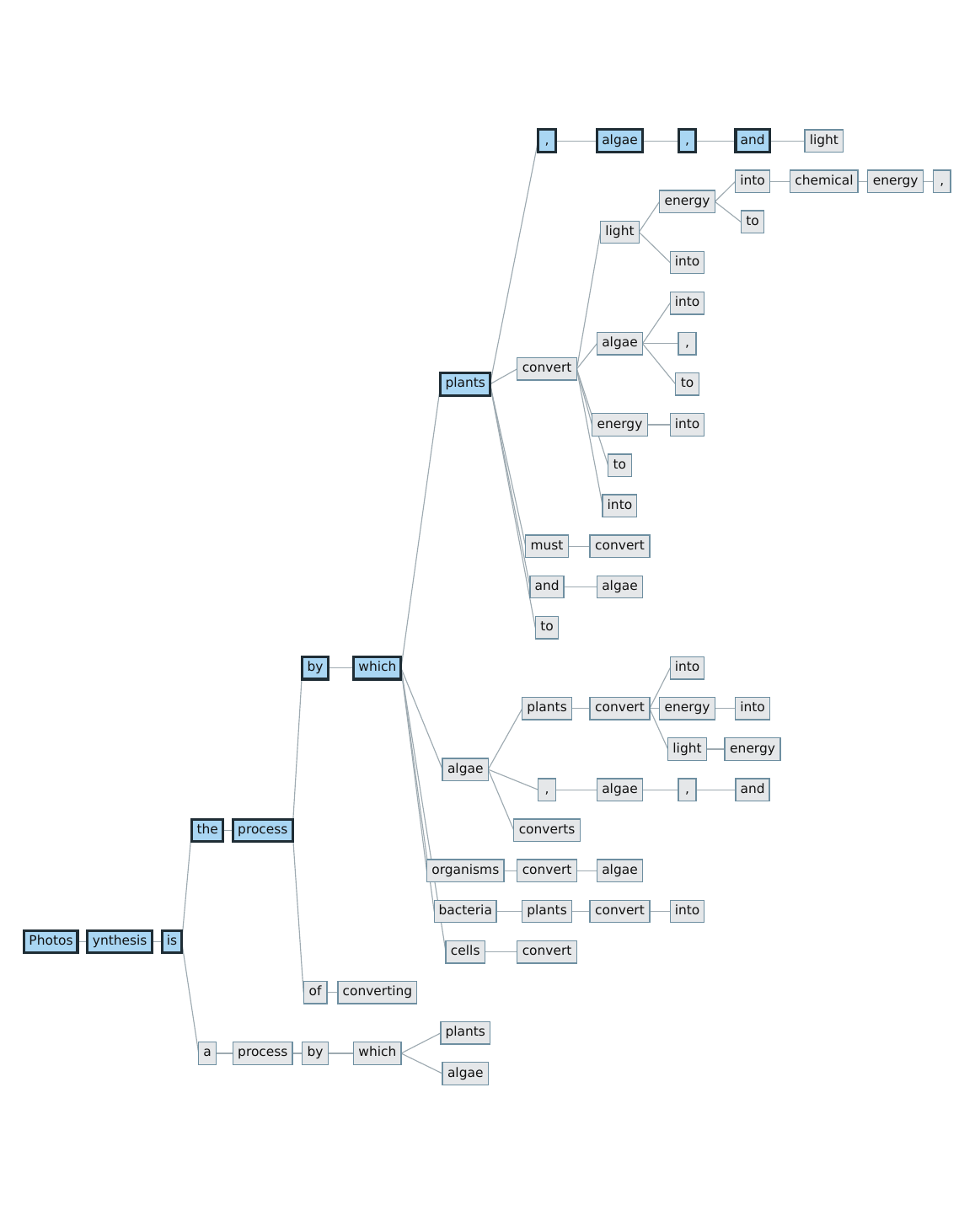}
    \caption{\textbf{DARTree verification tree for the photosynthesis prompt.}
    The prompt is
    ``\texttt{What is photosynthesis? Answer in exactly two short sentences.}''
    The tree begins at the first generated token. The target model accepts the twelve-token
    continuation
    \texttt{Photosynthesis is the process by which plants, algae, and}
    (\(\tau=12\)). Blue nodes with thick outlines mark the accepted path; gray
    nodes with thin outlines mark the remaining candidates.}
    \label{fig:dartree-photosynthesis-tree}
\end{figure*}

Figures~\ref{fig:ddtree-largest-animal-tree}--\ref{fig:dartree-photosynthesis-tree} visualize verification trees generated with Qwen3-4B at \(T=0\) for the prompts ``Can you tell me what is the largest animal in the world?'' and ``What is photosynthesis? Answer in exactly two short sentences.'' DDTree contains many repetitive or semantically incompatible parent--child transitions, such as ``It can about up'' and ``plants plants sunlight,'' whereas DARTree forms more coherent and well-structured branches. DARTree also yields longer accepted paths in both examples.

\subsection*{E. Implementation Details}
We provide additional implementation details of DARTree. Following prior work, the experiments in the main paper use the HuggingFace Transformers backend. For the concurrency evaluation in Table~\ref{tab:concurrency}, AR, DFlash, and Domino use their official SGLang implementations, while we implement DDTree and DARTree within the same SGLang framework for a fair comparison. We use Triton to accelerate several performance-critical operations, including candidate selection and tree-mask construction. As requested in the reproducibility checklist, we additionally report our system and hardware configuration. All experiments were conducted on a server running Ubuntu 22.04.1 LTS with two AMD EPYC 9374F CPUs, with each experiment using a single NVIDIA RTX 6000 Ada Generation GPU.

\subsection*{F. Future Work}
As shown in Figure~5 of the main paper, on some benchmarks, a substantial fraction of decoding rounds reach the end of the available draft block. This concentration near the drafting boundary suggests that the current block length may truncate continuations that could otherwise be accepted. A promising direction is therefore to train drafter models that support longer draft blocks and investigate whether DARTree can translate the expanded drafting horizon into higher acceptance and end-to-end speedups.

Another promising direction is to make the verification budget and tree shape adaptive rather than fixed across requests. These configurations could be selected according to runtime conditions, request load, and content-dependent signals such as draft confidence. Such adaptation may provide a better balance between acceptance gains and drafting and verification overhead across different deployment settings.

\clearpage

\clearpage

\clearpage

\clearpage


\end{document}